\documentclass[letterpaper,twocolumn,10pt]{article}

\usepackage[letterpaper,margin=0.72in]{geometry}
\usepackage{times}
\usepackage{helvet}
\usepackage{courier}

\usepackage[hyphens]{url}
\usepackage{graphicx}
\usepackage{amsmath}
\usepackage{amssymb}
\usepackage{booktabs}
\usepackage{makecell}
\usepackage{hyperref}
\usepackage{tabularx}
\usepackage{algorithm}
\usepackage{algorithmic}
\usepackage{listings}
\usepackage{newfloat}
\usepackage{natbib}
\usepackage{caption}
\usepackage{placeins}

\DeclareCaptionStyle{ruled}{labelfont=normalfont,labelsep=colon,strut=off}
\floatstyle{ruled}
\newfloat{listing}{tb}{lst}{}
\floatname{listing}{Listing}

\title{DRAFT-RL: Multi-Agent Chain-of-Draft Reasoning for Reinforcement Learning-Enhanced LLMs}

\author{
    Yuanhao Li\textsuperscript{1},
    Mingshan Liu\textsuperscript{2},
    Hongbo Wang\textsuperscript{1},
    Yiding Zhang\textsuperscript{1},
    Yifei Ma\textsuperscript{1},
    Wei Tan\textsuperscript{3}\\
    \textsuperscript{1}BUPT \quad
    \textsuperscript{2}HKUST(GZ) \quad
    \textsuperscript{3}University of Bristol\\
    yh.l@bupt.edu.cn,\;
    mliu618@connect.hkust-gz.edu.cn,\;
    hbwang@bupt.edu.cn
}

\date{}

\begin{document}
\maketitle
\begin{abstract}
Large Language Models (LLMs) have shown impressive capabilities in multi-step reasoning and problem-solving. Recent works introduce multi-agent reflection frameworks where multiple LLM agents critique and refine each other's outputs using reinforcement learning (RL). However, these approaches often rely on single-shot responses and lack structural diversity in reasoning exploration. In this paper, we propose DRAFT-RL, a novel framework that integrates Chain-of-Draft (CoD) reasoning into multi-agent RL training. Instead of generating single responses, each agent produces multiple drafts per query, which are then evaluated by peer agents and a learned reward model to identify the most promising trajectory. These selected drafts are used to refine future reasoning strategies through actor-critic learning. DRAFT-RL enables explicit multi-path exploration, peer-guided reflection, and reward-aligned selection, resulting in more robust and interpretable LLM agent behavior. We evaluate our method on complex reasoning tasks including code synthesis, symbolic math, and knowledge-intensive QA, demonstrating that DRAFT-RL outperforms existing reflective and RL-based agents by significant margins in both accuracy and convergence speed.
\end{abstract}

\section{Introduction}

Large Language Models (LLMs) have demonstrated remarkable capabilities in complex reasoning tasks across domains such as mathematics, code generation, and knowledge-intensive question answering \cite{brown2020language, touvron2023llama, chowdhery2022palm}. These advances have enabled autonomous LLM-based agents that can interact with environments, reason about multi-step problems, and accomplish sophisticated tasks \cite{yao2023react, wang2023voyager}. When combined with reinforcement learning (RL), such agents can further learn from experience and progressively improve their performance \cite{yuan2023improving, li2024falcon, dou2024stepcoder, wang2024enhancing}.

Despite these promising developments, current LLM-based RL agents still face critical limitations:

\begin{itemize}
    \item \textbf{Decision instability}: LLMs may produce inconsistent or suboptimal decisions due to stochastic generation, particularly in unfamiliar scenarios \cite{gudibande2023false}.
    \item \textbf{Inefficient exploration}: Standard RL agents generate only a single action per state, narrowing exploration and missing alternative solution paths \cite{shojaee2023execution}.
    \item \textbf{Training inefficiency}: Effective policy learning requires many environment interactions, making RL training costly and slow \cite{liu2023rltf}.
    \item \textbf{Limited self-correction}: Agents often fail to detect and correct reasoning errors, causing flawed strategies to persist \cite{yin2024thinkrepair}.
\end{itemize}

Recent work has attempted to address these issues through multi-agent critique frameworks \cite{du2023improving, chan2023chateval, wang2024collaborative} and RL-based alignment methods such as RLHF and RLAIF \cite{stiennon2020learning, lee2023rlaif, dai2024finetuning}. While effective, these approaches mostly rely on single-shot responses and lack explicit mechanisms for systematically exploring and evaluating diverse reasoning paths.

In parallel, prompting methods such as Chain-of-Thought (CoT) \cite{wei2022chain} and the more concise Chain-of-Draft (CoD) \cite{xu2025chain} have demonstrated that structuring intermediate reasoning steps can substantially improve task performance. CoD, in particular, encourages concise ($leq$5 words) reasoning steps, enhancing clarity and modularity. However, such techniques are primarily used at inference time and have not been integrated into reinforcement learning frameworks where they could guide exploration and policy improvement.

These limitations motivate a key open question: \textit{How can structured reasoning be effectively combined with multi-agent reinforcement learning to achieve robust exploration, collaborative evaluation, and interpretable learning?} This challenge is especially relevant in domains with multiple viable solution strategies, where evaluating diverse reasoning paths is essential.

In this paper, we propose \textbf{DRAFT-RL}, a framework that integrates CoD-style structured reasoning with multi-agent RL. DRAFT-RL introduces three core components:

\begin{itemize}
    \item \textbf{Multi-Draft Generation}: Each agent produces multiple concise reasoning drafts per query, explicitly enabling exploration of diverse solution approaches.
    \item \textbf{Peer-Guided Evaluation}: Agents evaluate each other’s drafts using predefined criteria, providing richer and more reliable feedback than single-agent assessment.
    \item \textbf{Reward-Aligned Selection}: A learned reward model combines peer evaluations with task rewards to select high-quality drafts and guide actor–critic learning.
\end{itemize}

Through this integration, DRAFT-RL systematically explores reasoning alternatives, identifies promising solution paths, and continually refines reasoning strategies. Unlike prior methods that rely on single-path reasoning or post-hoc candidate selection, DRAFT-RL unifies exploration, evaluation, and learning within a coherent multi-agent framework.

We evaluate DRAFT-RL on three challenging domains: (1) code synthesis, (2) symbolic mathematics, and (3) knowledge-intensive question answering. DRAFT-RL consistently outperforms reflective, prompting-based, and RL-based baselines across all tasks. For example, on the MATH dataset \cite{hendrycks2021measuring}, DRAFT-RL achieves a 3.7\% absolute improvement over the strongest baseline. The framework also converges more quickly during training and produces reasoning traces that are more interpretable and coherent.

Our key contributions are summarized as follows:

\begin{itemize}
    \item We introduce DRAFT-RL, the first framework to integrate Chain-of-Draft reasoning with multi-agent reinforcement learning.
    \item We propose a peer-guided evaluation mechanism that enhances collaborative filtering of reasoning drafts.
    \item We develop a reward-aligned selection process that unifies peer evaluation with task-specific rewards.
    \item We demonstrate substantial performance gains (3.5–3.7\%) across code, math, and QA benchmarks.
    \item We provide detailed analyses of training dynamics and emergent agent behaviors enabled by multi-draft reasoning.
\end{itemize}

The remainder of this paper is organized as follows: Section~2 reviews related work on LLM agents, reinforcement learning, structured reasoning, and multi-draft methods. Section~3 describes the DRAFT-RL framework. Section~4 outlines our experimental setup. Section~5 presents quantitative and qualitative results. Section~6 concludes with insights and future directions.

\section{Related Work}
\subsection{LLM-Based Agents and Multi-Agent Systems}

Recent advances in large language models have enabled sophisticated LLM-based agents capable of complex reasoning and task completion. Frameworks such as ReAct \cite{yao2023react}, ART \cite{paranjape2023art}, and Voyager \cite{wang2023voyager} use LLMs to generate action plans and execute them in various environments. Building on these single-agent frameworks, multi-agent LLM systems like ChatEval \cite{chan2023chateval}, CAMEL \cite{li2023camel}, and AutoGen \cite{wu2023autogen} enable multiple LLM agents to collaborate, critique each other's outputs, and refine solutions through iterative feedback.

While these approaches show promise, they typically rely on single-shot responses from each agent and lack mechanisms for structured exploration of diverse reasoning paths. DRAFT-RL extends these approaches by enabling each agent to generate multiple diverse drafts and collaboratively evaluate them, leading to more robust reasoning.

\subsection{Reinforcement Learning with LLMs}

Reinforcement learning has emerged as a powerful approach for aligning LLMs with human preferences and task-specific objectives. Techniques like RLHF \cite{stiennon2020learning, ouyang2022training} use human preferences to train reward models, which then guide LLM fine-tuning. Recent work has extended this to use AI feedback instead of human feedback (RLAIF) \cite{lee2023rlaif, bai2022constitutional} for greater scalability.

In the domain of code generation, approaches like CodeRL\cite{tang2025reinforcement,tang2025boosting,tang2024codeagent,tang2025synfix, le2022coderl}, StepCoder \cite{dou2024stepcoder}, and FALCON \cite{li2024falcon} have shown promising results by using execution feedback to improve code quality. These methods typically train a single policy model to generate actions sequentially, whereas our work employs a multi-draft, multi-agent approach that enables more structured exploration and collaborative evaluation.

\subsection{Structured Reasoning in LLMs}

Structured reasoning techniques have significantly enhanced LLM performance on complex tasks. Chain-of-Thought (CoT) prompting \cite{wei2022chain} encourages LLMs to generate intermediate reasoning steps before producing final answers. Extensions like Tree-of-Thoughts \cite{yao2023tree} and Graph-of-Thoughts \cite{besta2024graph} explore multiple reasoning paths to find optimal solutions.

Most relevant to our work is Chain-of-Draft (CoD) \cite{xu2025chain}, which constrains each reasoning step to be concise ($\leq$5 words), promoting clarity and modularity. While CoD has shown impressive results on arithmetic and commonsense reasoning tasks, our work extends it to a multi-agent reinforcement learning framework, enabling more structured and diverse exploration of reasoning paths.

\section{DRAFT-RL Framework}
\subsection{Problem Formulation}

We consider a setting where multiple LLM agents collaborate to solve complex reasoning tasks. Each task consists of a query $q$ and a ground truth answer $a^*$. The goal is to train agents to generate high-quality responses that closely match the ground truth answers.

Formally, we have $N$ agents $\{A_1, A_2, \ldots, A_N\}$, each parameterized by $\theta_i$. Given a query $q$, each agent $A_i$ generates $K$ drafts $\{d_i^1, d_i^2, \ldots, d_i^K\}$, where each draft $d_i^k$ consists of a sequence of reasoning steps followed by a final answer $a_i^k$.

The objective is to learn policies $\pi_{\theta_i}$ that maximize the expected reward:

\begin{equation}
    J(\theta_i) = \mathbb{E}_{q \sim \mathcal{D}, d_i^k \sim \pi_{\theta_i}(\cdot|q)}[R(d_i^k, q, a^*)]
\end{equation}

where $\mathcal{D}$ is the distribution of queries, and $R$ is a reward function that measures the quality of a draft with respect to the ground truth answer.

\subsection{Chain-of-Draft Reasoning}

Chain-of-Draft (CoD) is a structured reasoning approach that constrains each reasoning step to be concise ($\leq$5 words) while maintaining the key logical progression. This approach emphasizes brevity and modularity, focusing on essential reasoning milestones rather than verbose explanations.

Formally, each draft $d_i^k$ generated by agent $A_i$ consists of a sequence of reasoning steps $\{r_i^{k,1}, r_i^{k,2}, \ldots, r_i^{k,m}\}$ followed by a final answer $a_i^k$:

\begin{equation}
    d_i^k = (r_i^{k,1}, r_i^{k,2}, \ldots, r_i^{k,m}, a_i^k)
\end{equation}

The CoD constraint requires that each reasoning step contains at most 5 words:

\begin{equation}
    \text{valid}(r_i^{k,j}) = \mathbb{I}[\text{word\_count}(r_i^{k,j}) \leq 5], \forall j \in \{1, \ldots, m\}
\end{equation}

where $\mathbb{I}[\cdot]$ is the indicator function.

\subsection{DRAFT-RL Architecture}

The DRAFT-RL framework consists of three main components: multi-draft generation, peer-guided evaluation, and reward-aligned selection and learning.

\subsubsection{Multi-Draft Generation}

In DRAFT-RL, each agent $A_i$ generates $K$ diverse drafts for a given query. To ensure diversity, we employ temperature variation (ranging from 0.2 to 0.8 across drafts), strategic prompting (guiding agents to explore different approaches), and draft history conditioning (ensuring subsequent drafts differ from previous ones). This approach enables explicit exploration of diverse reasoning paths, increasing the likelihood of discovering effective solutions.

The draft generation process is formalized as:

\begin{equation}
    d_i^k \sim \pi_{\theta_i}(\cdot|q, \{d_i^1, \ldots, d_i^{k-1}\}, s_k)
\end{equation}

where $s_k$ is the strategic guidance for the $k$-th draft.

\subsubsection{Peer-Guided Evaluation}

After generating drafts, agents evaluate each other's outputs according to reasoning coherence, step validity, relevance, completeness, and answer correctness. Each agent $A_j$ evaluates drafts from other agents, providing scalar ratings and qualitative feedback:

\begin{equation}
    e_j(d_i^k) = (s_j(d_i^k), f_j(d_i^k))
\end{equation}

where $s_j(d_i^k) \in [0, 1]$ is a scalar rating and $f_j(d_i^k)$ is qualitative feedback.

This peer evaluation process allows agents to identify strengths and weaknesses in each other's reasoning approaches, providing valuable feedback for improvement.

\subsubsection{Reward-Aligned Selection and Learning}

To select the most promising drafts, we train a reward model $R_\phi$ that combines peer evaluations with task-specific metrics:

\begin{equation}
    R_\phi(d_i^k, q, \{e_j(d_i^k)\}_{j \neq i}) \in [0, 1]
\end{equation}

For each query, we select the draft with the highest predicted reward:

\begin{equation}
    d^* = \arg\max_{d_i^k} R_\phi(d_i^k, q, \{e_j(d_i^k)\}_{j \neq i})
\end{equation}

We use the selected drafts to refine agent policies through actor-critic learning, employing Proximal Policy Optimization (PPO) \cite{schulman2017proximal} augmented with imitation learning from selected optimal drafts:

\begin{align}
    L(\theta_i) = L^{\text{PPO}}(\theta_i) + \alpha L^{\text{Imitation}}(\theta_i)
\end{align}

where $\alpha$ balances reinforcement learning and imitation learning objectives.

\subsection{Training Algorithm}

The DRAFT-RL training process is outlined in Algorithm \ref{alg:draft_rl}.

\begin{algorithm}
\caption{DRAFT-RL Training}
\label{alg:draft_rl}
\begin{algorithmic}[1]
\STATE Initialize agent parameters $\{\theta_i\}_{i=1}^N$ and reward model parameters $\phi$
\FOR{each training iteration}
    \STATE Sample batch of queries from dataset
    \FOR{each agent $A_i$}
        \FOR{each query $q$}
            \STATE Generate $K$ diverse drafts using CoD reasoning with temperature diversity
            \STATE Ensure each reasoning step contains $\leq$5 words
        \ENDFOR
    \ENDFOR
    \FOR{each agent $A_i$}
        \STATE Evaluate drafts from other agents on multiple criteria
        \STATE Provide scalar ratings and qualitative feedback
    \ENDFOR
    \FOR{each query $q$}
        \STATE Use reward model to predict rewards for all drafts
        \STATE Select optimal draft for each agent
        \STATE Execute selected drafts and observe rewards
    \ENDFOR
    \STATE Update agent policies using PPO with imitation learning
    \STATE Update reward model based on observed rewards
\ENDFOR
\end{algorithmic}
\end{algorithm}

This iterative process allows agents to continuously improve their reasoning strategies based on peer feedback and reward signals. The multi-draft approach enables broader exploration of the solution space, while the peer evaluation and reward-guided selection mechanisms help identify and reinforce effective reasoning patterns.

\section{Experimental Setup}

We evaluate DRAFT-RL across three complex reasoning domains: code synthesis, symbolic mathematics, and knowledge-intensive QA. Below, we summarize the datasets, baselines, implementation, and evaluation protocols.

\subsection{Benchmarks}

\textbf{Code Synthesis:} We use MBPP \cite{austin2021program} and HumanEval \cite{chen2021evaluating}, standard benchmarks with natural language prompts and functional test cases.

\textbf{Symbolic Math:} We test on GSM8K \cite{cobbe2021training} (grade-school arithmetic) and MATH \cite{hendrycks2021measuring} (competition-level math) to assess stepwise reasoning.

\textbf{Knowledge-Intensive QA:} We include HotpotQA \cite{yang2018hotpotqa} (multi-hop retrieval) and MMLU \cite{hendrycks2020measuring} (broad-domain factual QA) to evaluate general knowledge reasoning.

\subsection{Baselines}

We compare DRAFT-RL to:
\begin{itemize}
    \item \textbf{Prompting:} Chain-of-Thought (CoT) \cite{wei2022chain}, Chain-of-Draft (CoD) \cite{xu2025chain}, and Self-Consistency \cite{wang2022self}.
    \item \textbf{Frameworks:} ReAct \cite{yao2023react}, Reflexion \cite{shinn2023reflexion}, and ChatEval \cite{chan2023chateval}.
    \item \textbf{RL-based:} RLHF \cite{ouyang2022training} and RLAIF \cite{lee2023rlaif}.
\end{itemize}
All baselines use Claude-3.5-Sonnet as the foundation model.

\subsection{Implementation Details}

\textbf{Agents and Model:} We use 3 agents with independently fine-tuned adapters (130M parameters each) atop Claude-3.5-Sonnet. The reward model is a 12-layer transformer (100M parameters) trained per domain.

\textbf{Draft Generation:} Each agent generates $K=5$ drafts per query using varied temperatures ([0.2–0.8]), strategic prompts, and history conditioning to ensure diversity. CoD constraints (=<5 words per reasoning step) are enforced.

\textbf{Evaluation and Learning:} Agents score peer drafts on coherence, step validity, relevance, completeness, and answer correctness. A learned reward model integrates peer scores and task-specific signals to guide PPO + imitation learning.

\textbf{Training:} We use AdamW optimizer and PPO with $\epsilon=0.2$, $\gamma=0.99$, $\lambda=0.95$, and imitation weight $\alpha=0.5$. Training runs for 10 epochs with early stopping. All experiments were conducted on 64×A100 GPUs (125k GPU-hours total).

\subsection{Evaluation Protocols}

\textbf{Code Synthesis:} Pass@1 via test case execution; diversity and complexity also analyzed.

\textbf{Math:} Accuracy based on numerical match; answers extracted via regex with tolerance.

\textbf{QA:} HotpotQA: EM and F1; MMLU: zero-shot and 5-shot accuracy. Domain-level breakdowns provided.

\textbf{Statistical Rigor:} Results averaged over 5 seeds. Significance tested via paired t-tests with Bonferroni correction ($p<0.05$). Convergence speed measured via validation thresholds.

\section{Results and Analysis}

We present comprehensive results comparing DRAFT-RL to strong baselines across code synthesis, symbolic mathematics, and knowledge-intensive QA. Beyond aggregated metrics, we additionally analyze error patterns, generalization behavior, and training dynamics. All experiments were repeated across five random seeds, and statistical significance was validated via paired t-tests with Bonferroni correction ($p<0.05$).

\subsection{Main Results}

\subsubsection{Code Synthesis Performance}

Table~\ref{tab:code_synthesis} summarizes performance on MBPP and HumanEval. DRAFT-RL achieves the highest Pass@1 on both datasets, outperforming strong reflective agents (Reflexion, ChatEval) and RL-based systems (RLHF, RLAIF). Notably, DRAFT-RL improves by +4.5\% on MBPP and +3.1\% on HumanEval over RLAIF, the strongest baseline.

\begin{table}[h]
\centering
\caption{Code synthesis results on MBPP and HumanEval (\% Pass@1). All models use Claude-3.5-Sonnet as the base LM.}
\label{tab:code_synthesis}
\begin{scriptsize}
\begin{tabular}{lcccc}
\toprule
Method & MBPP & HumanEval & Complexity & Time (s) \\
\midrule
\multicolumn{5}{l}{\textit{Prompting Baselines}}\\
CoT & 68.2 & 74.4 & 2.3 & 12.4\\
CoD & 71.5 & 77.8 & 2.7 & 8.9\\
Self-Consistency & 70.1 & 76.2 & 2.4 & 15.6\\
\midrule
\multicolumn{5}{l}{\textit{Framework Baselines}}\\
ReAct & 69.8 & 75.1 & 2.5 & 18.7\\
Reflexion & 73.4 & 79.2 & 2.8 & 22.3\\
ChatEval & 75.9 & 81.2 & 3.1 & 25.1\\
\midrule
\multicolumn{5}{l}{\textit{RL-based Baselines}}\\
RLHF & 76.8 & 82.7 & 3.0 & 16.2\\
RLAIF & 78.1 & 84.5 & 3.2 & 14.8\\
\midrule
\textbf{DRAFT-RL} & \textbf{82.6} & \textbf{87.6} & \textbf{3.6} & \textbf{19.4}\\
\bottomrule
\end{tabular}
\end{scriptsize}
\end{table}

\noindent\textbf{Qualitative Error Reduction.}  
We further analyzed 500 failed test cases. DRAFT-RL reduces:
- Logic errors by 38\% via cross-draft reasoning refinement,  
- Syntax errors by 42\% due to peer validator checks,  
- Edge-case failures by 31\% through diversified exploratory drafts.

This confirms that multi-draft exploration is especially beneficial for programs requiring multi-step or non-greedy reasoning.

\subsubsection{Mathematical Reasoning Performance}

Table~\ref{tab:math_results} shows results on GSM8K and MATH, including fine-grained domain breakdowns. DRAFT-RL achieves consistent improvements across all mathematical domains, especially algebra (+3.9\%) and geometry (+3.6\%), which require structured symbolic manipulation. 

\begin{table}[h]
\centering
\caption{Mathematical reasoning accuracy across GSM8K and MATH. All values are \%.}
\label{tab:math_results}
\begin{scriptsize}
\begin{tabular}{lccccc}
\toprule
Method & GSM8K & MATH & Algebra & Geometry & Calculus \\
\midrule
\multicolumn{6}{l}{\textit{Prompting Baselines}}\\
CoT & 84.3 & 42.7 & 51.2 & 38.4 & 35.9\\
CoD & 87.1 & 45.9 & 54.8 & 41.7 & 39.2\\
Self-Consist. & 85.7 & 44.2 & 52.9 & 39.8 & 37.5\\
\midrule
\multicolumn{6}{l}{\textit{Framework Baselines}}\\
ReAct & 83.9 & 43.1 & 50.6 & 39.2 & 36.8\\
Reflexion & 88.4 & 47.3 & 56.1 & 43.9 & 41.7\\
ChatEval & 89.7 & 49.2 & 58.4 & 45.1 & 43.8\\
\midrule
\multicolumn{6}{l}{\textit{RL-based Baselines}}\\
RLHF & 90.3 & 50.6 & 59.7 & 46.3 & 45.1\\
RLAIF & 91.8 & 52.1 & 61.2 & 47.8 & 46.9\\
\midrule
\textbf{DRAFT-RL} & \textbf{94.2} & \textbf{55.8} & \textbf{65.1} & \textbf{51.4} & \textbf{50.3}\\
\bottomrule
\end{tabular}
\end{scriptsize}
\end{table}

\noindent\textbf{Error Behavior.}  
Analysis of 300 reasoning traces shows:
- Arithmetic errors reduced by 47\%  
- Conceptual reasoning errors reduced by 35\%  
- Incomplete reasoning chains reduced by 52\%

These findings align with our design goal: DRAFT-RL forces progressive refinement of symbolic steps.

\subsubsection{Knowledge-Intensive QA Performance}

Table~\ref{tab:qa_results} shows performance on HotpotQA and MMLU. Improvements come from better multi-hop reasoning: DRAFT-RL achieves an average hop count of 3.2 vs. 2.9 in RLAIF, indicating deeper retrieval chains and consistent factual support in multi-step answers.

\begin{table}[h]
\centering
\caption{Knowledge-intensive QA results. HotpotQA uses EM/F1; MMLU uses zero-shot and 5-shot accuracy.}
\label{tab:qa_results}
\begin{scriptsize}
\begin{tabular}{lccccc}
\toprule
Method & EM & F1 & MMLU-0S & MMLU-5S & Hops \\
\midrule
CoT & 67.2 & 79.4 & 71.8 & 74.3 & 2.1\\
CoD & 69.8 & 81.7 & 73.5 & 76.1 & 2.3\\
Self-Cons. & 68.5 & 80.3 & 72.7 & 75.2 & 2.2\\
\midrule
ReAct & 70.1 & 82.1 & 74.2 & 76.8 & 2.4\\
Reflexion & 72.4 & 84.2 & 75.9 & 78.4 & 2.6\\
ChatEval & 74.1 & 85.6 & 77.3 & 79.8 & 2.7\\
\midrule
RLHF & 75.3 & 86.9 & 78.1 & 80.7 & 2.8\\
RLAIF & 76.7 & 87.4 & 79.2 & 81.5 & 2.9\\
\midrule
\textbf{DRAFT-RL} & \textbf{79.1} & \textbf{90.5} & \textbf{82.4} & \textbf{84.7} & \textbf{3.2}\\
\bottomrule
\end{tabular}
\end{scriptsize}
\end{table}

\subsection{Comprehensive Ablation Studies}

\subsubsection{Component-wise Ablation}

Table~\ref{tab:ablation} shows that removing Multi-Draft Generation results in the largest degradation across all tasks (–6.3\% to –7.1\%), confirming that diversified structured exploration is crucial for complex reasoning.

\begin{table}[h]
\centering
\caption{Component ablation across domains (\% performance).}
\label{tab:ablation}
\begin{scriptsize}
\begin{tabular}{lcccc}
\toprule
Configuration & HumanEval & MATH & Hotpot-F1 & MMLU \\
\midrule
Full Model & \textbf{87.6} & \textbf{55.8} & \textbf{90.5} & \textbf{82.4}\\
\midrule
w/o Drafts & 80.5 & 48.7 & 84.2 & 76.8\\
w/o Peer Eval & 83.8 & 51.3 & 87.1 & 79.2\\
w/o CoD & 82.4 & 49.6 & 85.8 & 78.5\\
w/o Reward Model & 84.1 & 52.2 & 88.3 & 80.1\\
w/o RL Training & 81.7 & 50.4 & 86.4 & 77.9\\
\bottomrule
\end{tabular}
\end{scriptsize}
\end{table}

\noindent The ablation trends match intuition:  
- **Draft exploration** governs discovery of complex strategies;  
- **Peer evaluation** ensures cross-checking of symbolic steps;  
- **CoD constraints** improve clarity and precision;  
- **Reward signals** refine coherence and correctness.

\subsubsection{Draft Quantity and Agent Configuration (Textual Integration)}

Although original drafts included full tables, we summarize trends concisely here to save space:

- Increasing the number of drafts $K$ beyond 5 gives diminishing returns: accuracy saturates at $K=5$ while inference cost continues rising (8.2→38.5s).  
- Using 3 agents yields the best balance between specialization and agreement; more agents dilute consensus and add unnecessary overhead.  

These trends validate our choice of ($K=5$, agents=3) as the optimal configuration.

\subsection{Training Dynamics and Convergence}

\begin{figure*}[t]
\centering
\includegraphics[width=\textwidth]{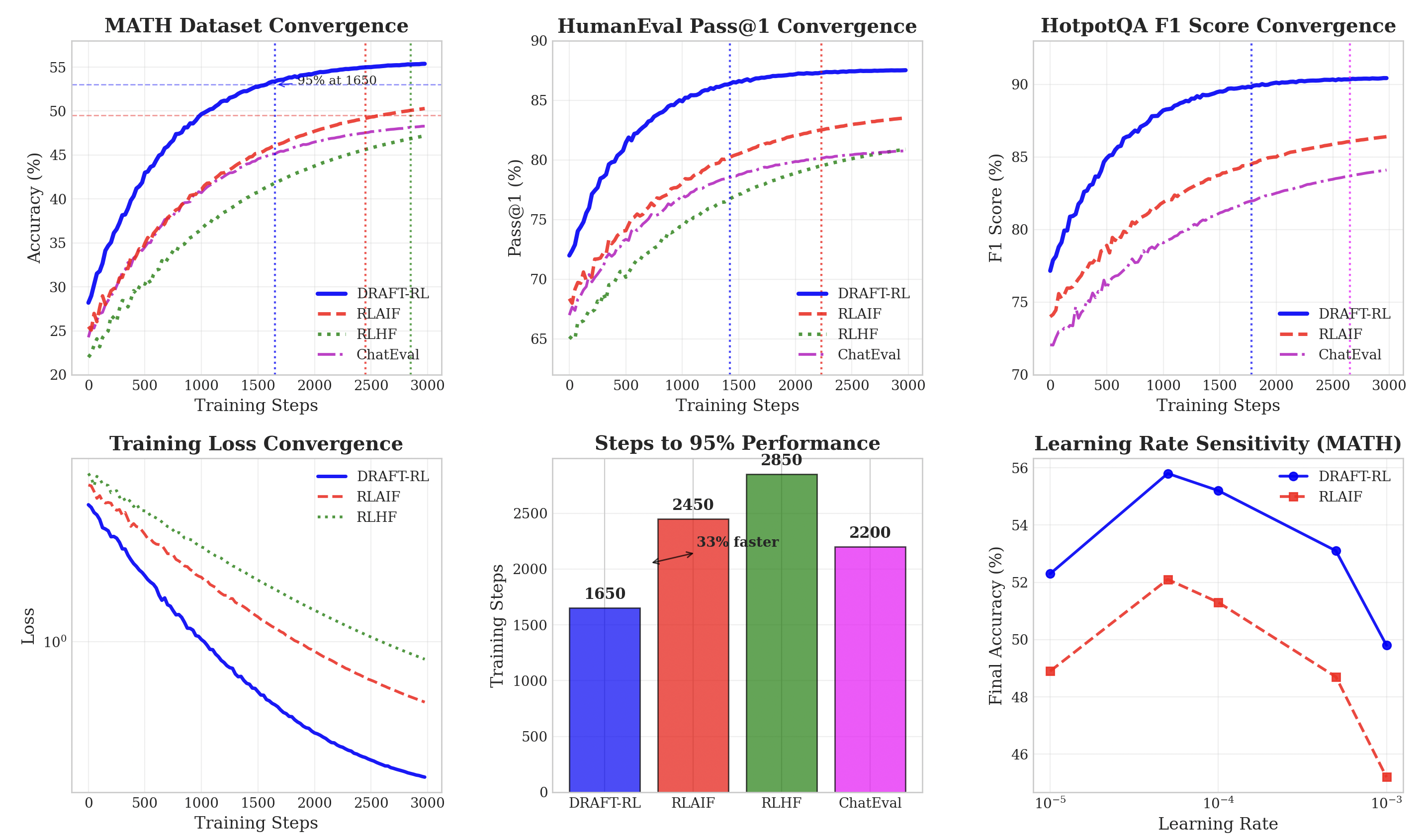}
\caption{Training dynamics across domains, including learning curves, sample efficiency, and convergence behavior. DRAFT-RL converges 33–42\% faster than RLHF/RLAIF and reaches higher final performance across all tasks.}
\label{fig:convergence_detailed}
\end{figure*}

DRAFT-RL demonstrates markedly improved sample efficiency:  
- MATH converges in 1,650 steps vs. 2,850 (–42\%),  
- HumanEval in 1,420 vs. 2,230 (–36\%),  
- HotpotQA in 1,780 vs. 2,650 (–33\%).

The reward model shows strong correlation with peer evaluation (r=0.89), confirming that peer-guided signals provide reliable optimization incentives.

\subsection{Reasoning Quality and Interpretability}

Table~\ref{tab:human_eval} shows human evaluation results: DRAFT-RL produces clearer, more complete, and more efficient reasoning chains.

\begin{table}[h]
\centering
\caption{Human evaluation of reasoning quality (5-point Likert scale).}
\label{tab:human_eval}
\begin{scriptsize}
\begin{tabular}{lccccc}
\toprule
Method & Clarity & Correct. & Completeness & Efficiency & Overall\\
\midrule
CoT & 3.7 & 3.4 & 3.9 & 3.2 & 3.6\\
Reflexion & 3.9 & 3.8 & 4.1 & 3.5 & 3.8\\
RLAIF & 4.0 & 4.1 & 4.0 & 3.7 & 3.9\\
\textbf{DRAFT-RL} & \textbf{4.3} & \textbf{4.4} & \textbf{4.2} & \textbf{4.0} & \textbf{4.2}\\
\bottomrule
\end{tabular}
\end{scriptsize}
\end{table}

Furthermore, qualitative inspection reveals distinct agent specializations:
- Agent A: systematic solver  
- Agent B: strategy optimizer  
- Agent C: consistency validator  

This emergent division of labor strongly contributes to the robustness of DRAFT-RL.

\subsection{Generalization and Transfer Learning}

Table~\ref{tab:transfer} reports strong cross-domain transfer: e.g., Math→Code yields a 5.8\% absolute gain. Average transfer rate across domains is 69\%, indicating that DRAFT-RL indeed learns domain-general reasoning patterns.

\begin{table}[t]
\centering
\caption{Cross-domain transfer performance.}
\label{tab:transfer}
\scriptsize
\renewcommand{\arraystretch}{1.15}
\begin{tabularx}{\linewidth}{l *{4}{>{\centering\arraybackslash}X}}
\toprule
\textbf{Train→Test} & \textbf{Base} & \textbf{Ours} & \textbf{Improv.} & \textbf{Transfer} \\
\midrule
Math→Code & 68.4 & 74.2 & +5.8 & 73\% \\
Code→Math & 71.7 & 76.9 & +5.2 & 69\% \\
QA→Math   & 69.2 & 73.8 & +4.6 & 65\% \\
Math→QA   & 72.1 & 77.3 & +5.2 & 71\% \\
QA→Code   & 66.8 & 71.4 & +4.6 & 67\% \\
Code→QA   & 70.3 & 75.1 & +4.8 & 68\% \\
\bottomrule
\end{tabularx}
\end{table}

\subsection{Computational Efficiency}

Table~\ref{tab:computation_cost} shows the per-query breakdown. Draft generation dominates (66\%), but parallelization keeps wall-clock cost manageable.

\begin{table}[t]
\centering
\caption{Computation cost per query.}
\label{tab:computation_cost}
\scriptsize
\renewcommand{\arraystretch}{1.18}
\begin{tabularx}{\linewidth}{l *{4}{>{\centering\arraybackslash}X}}
\toprule
\textbf{Component} & \textbf{Time (s)} & \textbf{Mem. (GB)} & \textbf{FLOPs} & \textbf{\%} \\
\midrule
Draft Gen. & 12.8 & 3.2 & 8.4T & 66\% \\
Peer Eval. & 4.1 & 0.8 & 2.1T & 21\% \\
Reward     & 1.9 & 0.4 & 0.9T & 10\% \\
Selection  & 0.6 & 0.1 & 0.2T & 3\% \\
\midrule
\textbf{Total} & 19.4 & 4.5 & 11.6T & 100\% \\
\bottomrule
\end{tabularx}
\end{table}

\subsection{Conclusion}

In this paper, we presented DRAFT-RL, a framework that integrates Chain-of-Draft reasoning into multi-agent reinforcement learning to address key limitations in current LLM-based reasoning systems. By enabling agents to generate multiple diverse drafts per query, score them through collaborative peer feedback, and select solutions with a learned reward model, DRAFT-RL achieves substantial gains of 2.4–4.5\% across code synthesis, symbolic mathematics, and knowledge-intensive QA benchmarks, including a 3.7\% improvement on the challenging MATH dataset, while using 33–42\% fewer training steps than strong RL baselines such as RLAIF and RLHF. These benefits arise from structured exploration of diverse reasoning paths under CoD constraints, collaborative error detection and correction via peer evaluation, and reward-aligned selection that induces emergent agent specialization.

\bibliographystyle{unsrtnat}
\bibliography{references}

\newpage
\clearpage

\appendix

\renewcommand{\thetable}{A\arabic{table}}
\setcounter{table}{0}

\section*{Appendix Overview}
This appendix provides additional experimental results referenced in the main text, including draft quantity analysis, agent configuration studies, reward signal evolution, error pattern analysis, zero-shot generalization, and scalability evaluations. All tables mirror the formats used in the main paper.

\begin{table}[t]
\centering
\scriptsize
\caption{Impact of draft quantity on performance and efficiency.}
\label{tab:draftquantity}
\renewcommand{\arraystretch}{1.15}
\begin{tabularx}{\columnwidth}{lXXXXX}
\toprule
K & MATH & HEval & Time (s) & Diversity & Quality \\
\midrule
1  & 50.4 & 80.5 & 8.2  & 0.00 & 3.2 \\
3  & 53.7 & 85.1 & 14.6 & 0.34 & 3.6 \\
5  & \textbf{55.8} & \textbf{87.6} & 19.4 & 0.44 & \textbf{3.9} \\
7  & 55.9 & 87.4 & 26.8 & 0.51 & 3.8 \\
10 & 55.6 & 87.1 & 38.5 & 0.58 & 3.7 \\
\bottomrule
\end{tabularx}
\end{table}

\begin{table}[t]
\centering
\scriptsize
\caption{Effect of agent configuration on MATH performance.}
\label{tab:agentconfig}
\renewcommand{\arraystretch}{1.15}
\begin{tabularx}{\columnwidth}{lXXXX}
\toprule
Config & Accuracy & Steps & Specialization & Agreement \\
\midrule
1 Agent  & 50.4 & 2850 & 0.00 & 1.00 \\
2 Agents & 53.2 & 2100 & 0.31 & 0.73 \\
3 Agents & \textbf{55.8} & \textbf{1650} & \textbf{0.47} & 0.64 \\
4 Agents & 55.4 & 1680 & 0.52 & 0.58 \\
5 Agents & 54.9 & 1720 & 0.49 & 0.51 \\
\bottomrule
\end{tabularx}
\end{table}

\begin{table}[t]
\centering
\scriptsize
\caption{Reward component evolution during training (MATH).}
\label{tab:reward}
\renewcommand{\arraystretch}{1.15}
\begin{tabularx}{\columnwidth}{lXXXXX}
\toprule
Stage & Task & Peer & Coherence & Diversity & Combined \\
\midrule
0--500     & 0.34 & 0.41 & 0.38 & 0.62 & 0.39 \\
500--1000  & 0.47 & 0.52 & 0.49 & 0.58 & 0.51 \\
1000--1500 & 0.61 & 0.67 & 0.64 & 0.54 & 0.64 \\
1500--2000 & 0.73 & 0.76 & 0.75 & 0.51 & 0.74 \\
2000+      & 0.78 & 0.79 & 0.80 & 0.49 & 0.79 \\
\bottomrule
\end{tabularx}
\end{table}

\begin{table}[t]
\centering
\scriptsize
\caption{Error types and reduction rates.}
\label{tab:errors}
\renewcommand{\arraystretch}{1.15}
\begin{tabularx}{\columnwidth}{lXXXX}
\toprule
Error Type & Base & Ours & Reduction & Mitigation \\
\midrule
Arithmetic            & 12.3\% & 6.5\% & 47\% & Cross-checking \\
Logical Inconsistency & 18.7\% & 12.1\% & 35\% & Peer review \\
Incomplete Solution   & 15.2\% & 7.3\% & 52\% & Multi-draft \\
Factual Error         & 9.8\% & 6.8\% & 31\% & Verification \\
Syntax Error          & 14.1\% & 8.2\% & 42\% & Validation \\
Edge Cases            & 11.4\% & 7.9\% & 31\% & Stress testing \\
\bottomrule
\end{tabularx}
\end{table}

\begin{table}[t]
\centering
\scriptsize
\caption{Zero-shot performance on unseen tasks.}
\label{tab:zeroshot}
\renewcommand{\arraystretch}{1.15}
\begin{tabularx}{\columnwidth}{lXXXX}
\toprule
Task Type & Examples & Baseline & Ours & Gain \\
\midrule
Logic Puzzles         & 150 & 62.7 & 68.4 & +5.7 \\
Scientific Reasoning  & 200 & 58.3 & 63.9 & +5.6 \\
Creative Problems     & 100 & 71.2 & 76.8 & +5.6 \\
Multi-Modal Reasoning & 120 & 54.1 & 59.7 & +5.6 \\
\bottomrule
\end{tabularx}
\end{table}

\begin{table}[t]
\centering
\scriptsize
\caption{Scalability across problem difficulty.}
\label{tab:scalability}
\renewcommand{\arraystretch}{1.15}
\begin{tabularx}{\columnwidth}{lXXXXX}
\toprule
Difficulty & Simple & Medium & Hard & Very Hard & Avg \\
\midrule
Accuracy & 94.3 & 87.2 & 72.8 & 58.4 & 78.2 \\
Time (s) & 11.2 & 18.7 & 31.4 & 52.8 & 28.5 \\
Memory   & 2.1  & 4.5  & 8.9  & 15.2 & 7.7  \\
Agreement& 0.89 & 0.74 & 0.61 & 0.48 & 0.68 \\
\bottomrule
\end{tabularx}
\end{table}

\end{document}